\pdfoutput=1

\documentclass[11pt,a4paper]{article}
\usepackage[hyperref]{acl2021}
\aclfinalcopy
\usepackage{times}
\usepackage{latexsym}

\usepackage[T1]{fontenc}
\usepackage[utf8]{inputenc}
\usepackage{microtype}

\usepackage{soul}
\definecolor{beaublue}{rgb}{0.74, 0.83, 0.9}
\definecolor{lavender}{rgb}{0.9, 0.9, 0.98}
\definecolor{tea}{rgb}{0.58, 0.87, 0.68}

\usepackage{booktabs}

\usepackage[linewidth=1pt]{mdframed}

\input{webis-paper-prefix}

\begin{document}
    \title{Claim Optimization in Computational Argumentation}

\author{  Gabriella Skitalinskaya\textsuperscript{1,2}, Maximilian Spliethöver\textsuperscript{1}, and Henning Wachsmuth\textsuperscript{1}\\
\textsuperscript{1}Leibniz University Hannover, Institute of Artificial Intelligence \\
 \textsuperscript{2}University of Bremen, Department of Computer Science \\
\small{\tt \{g.skitalinska,m.spliethoever,h.wachsmuth\}@ai.uni-hannover.de}\\ 
}
\maketitle
\date{}

\begin{abstract}
An optimal delivery of arguments is key to persuasion in any debate, both for humans and for AI systems. This requires the use of clear and fluent claims relevant to the given debate. Prior work has studied the automatic assessment of argument quality extensively. Yet, no approach actually improves the quality so far. To fill this gap, this paper proposes the task of \emph{claim optimization}: to rewrite argumentative claims in order to optimize their delivery. As multiple types of optimization are possible, we approach this task by first generating a diverse set of candidate claims using a large language model, such as BART, taking into account contextual information. Then, the best candidate is selected using various quality metrics. In automatic and human evaluation on an English-language corpus, our quality-based candidate selection outperforms several baselines, improving 60\% of all claims (worsening 16\% only). Follow-up analyses reveal that, beyond copy editing, our approach often specifies claims with details, whereas it adds less evidence than humans do. Moreover, its capabilities generalize well to other domains, such as instructional texts. 
\end{abstract}

    \interfootnotelinepenalty=10000

\section{Introduction}
\label{sec:introduction}

The delivery of arguments in clear and appropriate language is a decisive factor in achieving persuasion in any debating situation, known as {\em elocutio} in Aristotle's rhetoric \citep{elbaff:2019}. Accordingly, the claims composed in an argument should not only be grammatically fluent and relevant to the given debate topic, but also unambiguous, self-contained, and more.
Written arguments therefore often undergo multiple revisions in which various aspects are optimized \cite{zhang-litman-2015-annotation}.

Extensive research has been done on the automatic assessment of argument quality and the use of large language models on various text editing tasks. Yet, no work so far has studied how to actually improve argumentative texts.
However, developing respective approaches is a critical step towards building effective writing assistants, which could help learners write better argumentative texts \cite{wambsganss:2021} or rephrase arguments made by an AI debater \cite{slonim:2021}. In this work, we close the outlined gap by studying how to employ language models for rewriting argumentative text to optimize its delivery.

\begin{figure}[t]
	\centering
	\includegraphics[scale=1.0]{figures/example-revisions-2}
	\caption{Examples of different optimized versions of an \emph{original claim} found on the debate platform {Kialo}. All optimizations were generated by the approach proposed in this paper, using the \emph{debate topic} as context.}
	\label{example_revisions}
\end{figure}

We start by defining the task of \emph{claim optimization} in Section~\ref{sec:taxonomy}, and adjust the English-language claim revision dataset of \citet{skitalinskaya2021} for evaluation. The new task requires complementary abilities: On the one hand, different types of quality issues inside a claim must be detected, from grammatical errors to missing details. If not all quality aspects can be improved simultaneously, specific ones must be targeted. On the other hand, improved claim parts need to be integrated with the context of the surrounding discussion, while preserving the original meaning as far as possible. Figure~\ref{example_revisions} shows three exemplary optimizations of a claim from the debate platform {\em Kialo}. The first elaborates what the consequence of weaponization is, whereas the second rephrases the claim to clarify what weaponizing means, employing knowledge about the debate topic. The third renders the stance of the claim explicit. We observe that different ways to optimize a claim exist, yet the level of improvement differs as well.

To account for the multiplicity of claim optimization, we propose a controlled generation approach that combines the capabilities of large language models with quality assessment (Section~\ref{sec:approach}). First, a fine-tuned generation model produces several candidate optimizations of a given claim. To optimize claims, we condition the model on discourse context, namely the debate topic and the previous claim in the debate. The key to selecting the best optimi\-zation is to then score candidates~using three quality metrics:
 \emph{grammatical fluency}, \emph{meaning preservation}, and \emph{argument quality}. Such candidate selection remains understudied in many generative tasks, particularly within computational argumentation.

In automatic and manual evaluation (Section~\ref{sec:experiments}), we demonstrate the effectiveness of our approach, employing fine-tuned BART \cite{lewis-etal-2020-bart} for candidate generation.
Our results stress the benefits of quality assessment (Section~\ref{sec:results}). Incorporating context turns out especially helpful for making shorter claims---where the topic of the debate is difficult to infer---more self-contained. According to human annotators, our approach improves~60\% of all claims and harms only~16\%, clearly outperforming standard fine-tuned generation.

To gain further insights, we carry out a manual annotation of 600 claim optimizations and identify eight types typically found in online debate communities, such as {\em elaboration} and {\em disambiguation} (Section~\ref{sec:analysis}). Intriguingly, our approach covers similar optimization types as in human revisions, but we also observe limitations (Section~\ref{sec:analysis}). To explore to what extent it generalizes to other revision domains, we also carry out experiments on instructional texts \cite{anthonio-roth-2020-learn} and formal texts \cite{du-etal-2022-understanding-iterative}, finding that it outperforms strong baselines and state-of-the-art approaches.

In summary, the contributions of this paper are:
\begin{enumerate}
\setlength{\itemsep}{-2pt}
\item
\emph{a new task}, claim optimization, along with a manual analysis of typical optimization types;
\item
\emph{a computational approach} that selects the best generated candidate claim in terms of quality;
\item
\emph{empirical insights} into the impact and challenges of optimizing claims computationally.%
\footnote{Data, code, and models from our experiments are found at \href{https://github.com/GabriellaSky/claim_optimization}{https://github.com/GabriellaSky/claim\_optimization}}
\end{enumerate}
    \section{Related Work}
\label{sec:relatedwork}

Quality assessment has become a key topic in computational argumentation research \cite{lapesa:2023}. Various quality dimensions exist in argumentation theory, as surveyed by \newcite{wachsmuth-etal-2017-computat} and assessed computationally in various works \cite{lauscher:2020,marro:2022}. Many of them relate to quality aspects we consider in this work, from clarity and organization \cite{wachsmuth:2016} to the general evaluability of arguments \cite{park:2018}, potential fallacies in their reasoning \cite{goffredo:2022}, and the appropriateness of the language used \cite{ziegenbein:2023}. Recently, \cite{skitalinskaya:2023} tackled the question whether an argumentative claim is in need of revision, whereas \newcite{jundi:2023} investigated where to best elaborate a discussion. While \newcite{gurcke:2021} leverage claim generation for a refined assessment of argument quality, we are not aware of any prior work that actually optimizes arguments or their components in order to improve quality.

As shown in Figure~\ref{example_revisions}, there can be several ways to optimize a given text. Our key idea is to select the best optimization among diverse candidates generated by a language model. Prior generation work on candidate selection hints at the potential benefits of such setup, albeit in other tasks and domains. In early work on rule-based conversational systems, \citet{walker-etal-2001-quantitative} introduced dialogue quality metrics to optimize template-based systems towards user satisfaction. \citet{kondadadi-etal-2013-statistical} and \citet{cao-etal-2018-retrieve} chose the best templates for generation, and \citet{mizumoto-matsumoto-2016-discriminative} used syntactic features to rank candidates in grammar correction. Recently, \citet{yoshimura-etal-2020-reference} proposed a reference-less metric trained on manual evaluations of grammar correction system outputs to assess generated candidates, while \citet{suzgun2022promtandrerank} utilize pre-trained language models to select the best candidate in textual style transfer tasks. 

In generation research on computational argumentation, candidate selection remains largely understudied. Most relevant in this regard is the approach of \citet{chakrabarty-etal-2021-entrust} which reframes arguments to be more trustworthy (e.g., less partisan). It generates multiple candidates and selects one based on the entailment relation scores to the input. Extending this idea, we select candidates based on various properties, including argument quality.

Understanding the editing process of arguments is crucial, as it reveals what quality dimensions are considered important. For Wikipedia, \citet{daxenberger-gurevych-2013-automatically} proposed a fine-grained taxonomy as a result of their multi-label edit categorization of revisions \cite{daxenberger-gurevych-2012-corpus}. The taxonomy focuses solely on the editing actions performed, such as inserting, deleting, and paraphrasing. In contrast, \citet{yang_identifying_2017} identified various semantic intentions behind Wikipedia revisions, from {\em copy editing} to {\em content clarifications} and {\em fact updates}. Their taxonomy defines a starting point for our research. Not all covered intentions generalize beyond Wiki scenarios, though.\,

Wikipedia-based corpora have often been used in the study of editing and rewriting, including paraphrasing \cite{max-wisniewski-2010-mining}, grammar correction \cite{lichtarge-etal-2019-corpora}, bias neutralization \cite{pryzant-etal-2020}, and controllable text editing \cite{faltings-etal-2021-text,du-etal-2022-understanding-iterative}.  Similarly, WikiHow enabled summarization \cite{koupaee-wang-2018} and knowledge acquisition \cite{zhou-etal-2019-learning-household}. However, neither of these includes {\em argumentative} texts. Instead, we thus rely on the corpus of \citet{skitalinskaya2021}, which consists of revision histories of argumentative claims from online debates. Whereas the authors \emph{compare} claims in terms of quality, we propose and study the new task of automatically \emph{optimizing} claim quality.
Moreoever, we see the revision types they distinguish (clarification, grammar correction, linking to external sources) as too coarse-grained to represent the diversity of claim optimizations. We refine them manually into eight optimization types, allowing for a more systematic analysis. 
\citet{skitalinskaya2021} also found low correlations between the revision types and 15 common argument quality dimensions \cite{wachsmuth-etal-2017-computat}, suggesting that they are rather complementary. Primarily, they target the general form a well-phrased claim should have and its relevance to the debate.

For the analysis of argumentative text rewriting, \citet{zhang-litman-2015-annotation} incorporated both argumentative writing features and surface changes. To explore the classification of essay revisions, they defined a two-dimensional schema, combining the revision operation (e.g., modify, add, or delete) with the component being revised (e.g., reasoning or evidence). Moreover, \citet{afrin-litman-2018-annotation} created a small corpus of between-draft revisions of 60 student essays to study whether revision improves quality. However, these works do not uncover the reasoning behind a revision operation and are more geared towards analysis at the essay level.

    \section{Task and Data}
\label{sec:taxonomy}

This section introduces the proposed task and pre-sents the data used for development and evaluation.

\subsection{Claim Optimization}

We define the claim optimization task as follows:

\paragraph{Task}
Given as input an argumentative claim $c$, potentially along with context information on the debate, rewrite $c$ into an output claim $\tilde{c}$ such that
\begin{itemize}
\setlength{\itemsep}{0pt}
\item[(a)]
$\tilde{c}$ improves upon $c$ in terms of text quality and/or argument quality, and
\item[(b)]
$\tilde{c}$ preserves the meaning of $c$ as far as possible.
\end{itemize}

While we conceptually assume that $c$ consists of one or more sentences and has at least one quality flaw, our approaches do not model this explicitly. Moreover, note that $c$ might have multiple flaws, resulting in $n \geq 2$ candidate optimizations $\tilde{C} = \{ \tilde{c}_1, \ldots, \tilde{c}_n \}$. In this case, the goal is to identify the candidate~$c^* \in \tilde{C}$ that maximizes overall quality.

\subsection{Data for Development and Evaluation}
\label{subsec:data}

We start from the ClaimRev dataset \cite{skitalinskaya2021}, consisting of 124,312 claim revision histories from the debate platform {\em Kialo}. Each history defines a chain $(c_1, ..., c_m)$, in which claim~$c_{i}$ is a revised version of the previous claim, $c_{i-1}$~with $1< i\leq~m$, improving upon its quality. According to the authors, this holds in 93\% of all cases.

From each revision chain, we derived all possible optimization pairs $(c, \tilde{c}) := (c_{i-1}, c_i)$, in total 210,222. Most revisions are labeled with their intention by the users who performed them, rendering them suitable for learning to optimize claims automatically.%
\footnote{As 26\% of all pairs were unlabeled, we trained a BERT model to assign such pairs one of the 6 most prominent labels.}
Overall, 95\% of all pairs refer to three intention labels: {\em clarification}, {\em typo/grammar correction}, and {\em corrected/added links}. To avoid noise from the few remaining labels, we condensed the data to 198,089 instances of the three main labels.%
\footnote{The labels of the removed instances denote changes to the meaning of $c$ and statements from which no action or intention can be derived (e.g., "see comments", "moved as pro").}

For the final task dataset, we associated each remaining pair  $(c, \tilde{c})$ to its context: the {\em debate topic} $\tau$ (i.e., the thesis on Kialo) as well as the {\em previous claim} $\hat{c}$ (the parent on Kialo), which is supported or opposed by $c$ (see Figure \ref{example_revisions}). We sampled 600 revision pairs pseudo-randomly as a test set (200 per intention label), and split remaining pairs into training (90\%) and validation set (10\%). As the given labels are rather coarse-grained, we look into the optimizations in more detail in Section~\ref{sec:analysis}.

    \section{Approach}
\label{sec:approach}

We now present the first approach to automatic claim optimization. To account for the variety of possible optimizations, multiple candidate claims are generated that are pertinent to the context given and preserve the claim's meaning. Then, the best candidate is selected based on quality metrics. Both steps are detailed below and illustrated in Figure~\ref{fig:approach}.

\subsection{Seq2Seq-based Candidate Generation}

To generate candidates, we fine-tune a Seq2Seq model on pairs $(c, \tilde{c})$, by treating the original claim $c$ as encoder source and revised claim $\tilde{c}$ as the decoder target. In a separate experiment, we condition the model on context information, the debate topic $\tau$ and the previous claim $\hat{c}$, during fine-tuning to further optimize the relevance of generated candidates. The context is separated from $c$ by delimiter tokens \cite{keskar-etal-2019-ctlr,schiller-etal-2021-aspect}.

Multiple ways to improve $c$ exist, especially if it suffers from multiple flaws, since not all flaws may be fixed in a single revision. Therefore, we first generate $n$ suitable candidates, $\tilde{c}_1, \ldots, \tilde{c}_n$, among which the best one is to be found later ($n$ is set to 10 in Section~\ref{sec:experiments}). However, the top candidates created by language models often tend to be very similar. To increase the diversity of candidates, we perform top-$k$ sampling \cite{fan-etal-2018-hierarchical}, where we first generate the most probable claim (top-$1$) and then vary $k$ with in steps of 5 (e.g. top-$5$, top-$10$, etc).

\begin{figure}[t]
	\centering
	\includegraphics[scale=0.995]{figures/approach.ai}
	\caption{Proposed claim optimization approach: First, we generate $n$ candidates from the {\em original claim}, possibly conditioned on context information. Then, the {\em optimized claim} is selected using three quality metrics.
	}
	\label{fig:approach}
\end{figure}

\subsection{Quality-based Candidate Selection}
\label{sec_method_optimization}

Among the $n$ candidates, we aim to find the optimal claim, $c^*$, that most improves the delivery of $c$ in terms of text and argument quality. Similar to \citet{yoshimura-etal-2020-reference}, we tackle this task as a candidate selection problem. In our proposed strategy, {\em AutoScore}, we integrate three metrics: (1)~grammatical fluency, (2)~meaning preservation, and (3)~argument quality. This way, we can \textit{explicitly} favor specific quality dimensions via respective models:

\paragraph{Grammatical Fluency}

We learn to assess fluency on the MSR corpus \cite{toutanova-etal-2016-dataset} where the grammaticality of abstractive compressions is scored by 3--5 annotators from 1 (disfluent) to 3 (fluent).
We chose this corpus, since multiple compressions per input make a trained model sensitive to the differences in variants of a text. For training, we average all annotator scores and make the task binary, namely, a text is seen as disfluent unless all annotators gave score~3. Then, we train BERT on the data to obtain fluency probabilities (details found in Appendix~\ref{app:implementation}). The accuracy of our model on the suggested data split is 77.4.

\paragraph{Meaning Preservation}

To quantify to what extent a generated candidate maintains the meaning of the original claim, we compute their semantic similarity as the cosine similarity of the SBERT sentence embeddings \cite{reimers:2019}.

\paragraph{Argument Quality}

Finally, to examine whether the generated candidates are better than the original claim from an argumentation perspective, we fine-tune a BERT model on the task of pairwise argument classification using the ClaimRev dataset. Since this corpus is also used to fine-tune the Seq2Seq model, we apply the same training and validation split as described in Section \ref{subsec:data} to avoid data leakage, and obtain 75.5 accuracy. We then use its probability scores to determine relative quality improvement (for more details see Appendix~\ref{app:implementation}).

\medskip
Given the three quality metrics, we calculate the final evaluation score, $AutoScore$, as the weighted linear sum of all three individual scores as
$$
\alpha \cdot fluency + \beta \cdot meaning + \gamma \cdot argument,
$$
where $fluency$, $meaning$, and $argument$ are normalized scores of the three outlined quality metrics. The non-negative weights satisfy $\alpha+\beta+\gamma=1$.

\medskip
It should be noted that depending on the domain or writing skills of the users, there may be other more suitable datasets or approaches to capturing the outlined quality aspects, which could potentially lead to further performance improvements. While we do explore how well the suggested approaches transfer to certain other domains of text (see Section~\ref{subsec:domains}),  identifying the optimal model for each quality dimension falls beyond the scope of this paper.

    \section{Experiments}
\label{sec:experiments}

This section describes our experimental setup to study how well the claims from Section~\ref{sec:taxonomy} can be improved using our approach from Section~\ref{sec:approach}. We focus on the impact of candidate selection.

\subsection{Seq2Seq-based Candidate Generation}

For candidate generation, we employ the pre-trained conditional language model BART
 \cite{lewis-etal-2020-bart}, using the \textit{bart-large} checkpoint.
However, other Seq2Seq architectures can also be considered within our approach (see Appendices~\ref{app:implementation},~\ref{sec:alternatives}).

\subsection{Quality-based Candidate Selection}

We evaluate our candidate selection approach in comparison to three ablations and four baselines:

\paragraph{Approach}

To utilize AutoScore for choosing candidates, the optimal weighting of its metrics must be determined. We follow \citet{yoshimura-etal-2020-reference}, performing a grid search in increments of $0.01$ in the range of $0.01$ to $0.98$ for each weight to maximize the Pearson's correlation coefficient between AutoScore and the original order of the revisions from revision histories in the validation set. Similar has been done for counterargument retrieval by \citet{wachsmuth:2018}.
The best weights found are $\alpha\!=\!0.43, \beta\!=\!0.01$, and $\gamma\!=\!0.56$, suggesting that meaning preservation is of low importance and potentially may be omitted. We suppose this is due to the general similarity of the generated candidates, so a strong meaning deviation is unlikely.

\paragraph{Ablations}

To assess the impact of each considered quality metric used in AutoScore, we perform an ablation study, where optimal candidates are chosen based on the individual metric scores:
\begin{itemize}
\setlength{\itemsep}{-4pt}
    \item \emph{Max Fluency.} Highest grammatical fluency
    \item \emph{Max Argument.} Highest argument quality
    \item \emph{Max Meaning.} Highest semantic similarity
\end{itemize}

\paragraph{Baselines}

We test four selection strategies for 10~candidates generated via top-$k$ sampling:

\begin{itemize}
 \setlength{\itemsep}{-2pt}
    \item \emph{Unedited.} Return the original input as output.
    \item \emph{Top-1.} Return the most likely candidate (obtained by appending the most probable token generated by the model at each time step).
    \item \emph{Random.} Return candidate pseudo-randomly.
    \item \emph{SVMRank.} Rerank candidates with SVMRank
    \cite{joachims_svm2006}.\,Using sentence embeddings we decide which of the claim versions is better, by fine-tuning SBERT ({\em bert-base-cased})
    on the corpus of \citet{skitalinskaya2021}.
\end{itemize}

\subsection{Evaluation}

We explore claim optimization on all 600 test cases, both automatically and manually:

\paragraph{Automatic Evaluation}

We compare all content selection strategies against the reference revisions using the precision-oriented {\em BLEU} \cite{papineni-etal-2002-bleu}, recall-oriented {\em  Rouge-L} \cite{lin-2004-rouge}, {\em SARI} \cite{xu-etal-2016-optimizing}, which computes the average F$_1$-scores of the added, kept, and deleted $n$-grams in comparison to the ground truth revision output, and the \emph{exact match accuracy}.
We also compute the semantic similarity of the optimized claim and the context information to capture whether conditioning claims on context affects their topic relevance.

\paragraph{Manual Evaluation}

As we fine-tune existing generation models rather than proposing new ones, we focus on the {\em candidate selection} in two manual annotation studies. For each instance, we acquired five independent crowdworkers via {\em MTurk}.

In the first study, the annotators scored all candidates with respect to the three considered quality metrics. We used the following Likert scales:
 \begin{itemize}
 \setlength{\itemsep}{-2pt}
 \item
 \textit{Fluency.} 1 (major errors, disfluent), 2 (minor errors), and 3 (fluent)
 \item
 \textit{Meaning Preservation.} 1 (entirely different), 2 (substantial differences), 3 (moderate differences), 4 (minor differences), and 5 (identical)
 \item
 \textit{Argument Quality.} 1 (notably worse than original), 2 (slightly worse), 3 (same as original), 4 (slightly improved), and 5 (notably improved)
 \end{itemize}

A challenge of crowdsourcing is to ensure good results \cite{sabou-etal-2014-corpus}. To account for this, we obtained the fina fluency, argument quality and meaning preservation scores using MACE \cite{hovy-etal-2013-learning}, a Bayesian model that gives more weight to reliable workers. In the given case, 39\% of the 46 annotators had a MACE competence value $> 0.3$, which can be seen as reasonable in MTurk studies.

In the second study, we asked annotators to rank four candidates, returned by the content selection strategies, by perceived overall quality. If multiple candidates were identical, we showed each only once. While Krippendorff's~$\alpha$ agreement was only 0.20 and percent agreement was 0.36\% (majority voting), such values are common in subjective tasks \cite{wachsmuth-etal-2017-computat,alshomary-etal-2021-belief}.

\section{Results and Discussion}
\label{sec:results}
Apart from evaluating the applicability of large generative language models to the task of argumentative claim optimization in general, our experiments focus on two questions:\,(1)~Does the use of explicit knowledge about text and argument quality %in the decoding step
lead to the selection of better candidates?
(2)~Does the use of contextual information make the generated candidates more accurate and relevant to the debate?

\subsection{Overall Claim Optimization Performance}
\label{ranking_results}

\paragraph{Automatic Evaluation}

Table \ref{tab:exp1-auto} shows the automatic scores of all considered candidate selection strategies. The high scores of the baseline \emph{Unedited} on metrics such as BLEU and ROUGE-L indicate that many claim revisions change little only. In contrast, \emph{Unedited} is worst on SARI, a measure taking into account words that are added, deleted, and kept in changes, making it more suitable for evaluation. Here, \emph{BART+AutoScore} performs best on SARI (43.7) and exact match accuracy (8.3\%).

The \emph{BART+Max Meaning} ablation supports the intuition that the candidates with highest meaning preservation scores are those with minimal changes, if any (72\% of the candidates remain identical to the input). Such identical outputs are undesirable, as the claims are not optimized successfully, which is also corroborated by the low weight parameter ($\beta = 0.01$) found for the meaning preservation metric when optimizing AutoScore (see Section~\ref{sec:experiments}).

\begin{table}[t]
    \renewcommand{\arraystretch}{0.95}
    \setlength{\tabcolsep}{2pt}
    \centering
    \small
    \begin{tabular}{@{}l@{\hspace*{-0.75em}}rrrcr}
        \toprule
        \bf Approach &
        \multicolumn{1}{l}{\bf BLEU} &
        \multicolumn{1}{r}{\bf RouL} &
        \multicolumn{1}{l}{\bf SARI} &
        \multicolumn{1}{r}{\bf NoEd$\downarrow$} &
        \multicolumn{1}{c}{\bf ExM}  \\

        \midrule

\bf Baselines \\
 \, Unedited & \bf 69.4 & \bf 0.87 & 27.9 & 1.00 & 0.0\% \\

 \, BART + Top-1          		& 64.0	    &  0.83 		& 39.7	& 0.31    & 7.8\%     \\

  \, BART + Random     		& 62.6      & 0.83 		& 38.7 	& 0.28    & 6.8\%      	\\

  \,	BART + SVMRank    		& 55.7      & 0.76 		& 38.8  & 0.03  & 4.5\%        	\\[0.5ex]

\bf Approach  \\

        	 \,	BART + AutoScore		& 59.4      & 0.80 	    &  \bf 43.7 	& \textbf{0.02} & \bf 8.3\%	\\[0.5ex]

\bf Ablation \\
 \, BART + Max Fluency           &  57.6     & 0.78      & 41.5  & 0.09 & 5.8\% \\
        		 \,	BART + Max Argument          & 60.9      & 0.81      &  43.6  &  0.02 & 8.0\%\\
        		 \,	BART + Max Meaning           &  69.0  &  0.87      & 33.8 & 0.72 & 5.2\% \\
\bottomrule
\end{tabular}
\caption{Automatic evaluation: Performance of each candidate selection strategy on 600 test cases in terms~of~BLEU, Rouge-L, SARI, ratio of unedited cases, and ratio of exact matches to target reference.}
\label{tab:exp1-auto}
\end{table}

\begin{table}[t]
    \small
    \renewcommand{\arraystretch}{0.95}
    \setlength{\tabcolsep}{2.25pt}
    \centering
    \begin{tabular}{llrrrrcrrrr}
        \toprule
        \bf Model &
        \bf Strategy  & 
        \multicolumn{1}{l}{\bf Fluency} &
        \multicolumn{1}{l}{\bf Argument} &
        \multicolumn{1}{l}{\bf Meaning} &
       \cellcolor[rgb]{0.95, 0.95, 0.95} \bf Rank \\

        \midrule
        
BART		& Top-1          		& 2.29 		& 3.61 & 3.65  		&  \cellcolor[rgb]{0.95, 0.95, 0.95} 2.16 \\
& Random     		&  2.26 		& 3.50 		& 3.53  		& \cellcolor[rgb]{0.95, 0.95, 0.95} 2.06\\
& SVMRank    		&  \textbf{2.33} 		&  \textbf{3.69} 		& \bf 3.66  & \cellcolor[rgb]{0.95, 0.95, 0.95} 1.95 \\
& \bf AutoScore		&  \textbf{2.33} 	& 3.61  		&  3.57	& \cellcolor[rgb]{0.95, 0.95, 0.95} \textbf{1.92}\\
\bottomrule
\end{tabular}
\caption{Manual evaluation: Scores on the 600 test cases generated by BART using our candidate selection strategy \emph{AutoScore} or the baselines:  fluency (1--3), argument quality and meaning (1--5), mean rank (1--4, lower better). {AutoScore} ranks significantly better than \emph{Top-1} ($p<.005$), \emph{Random} ($p<.05$), and \emph{SVMRank} ($p<.1$).}
\label{tab:exp1-human}
\end{table}

\paragraph{Manual Evaluation}

Table~\ref{tab:exp1-human} shows that human annotators prefer optimized candidates selected by {\em AutoScore}, with an average rank of 1.92.
The difference to {\em Top-1} and {\em Random} is statistically significant ($p <.05$ in both cases) according to a Wilcoxon signed-rank test, whereas the gain over the second-best algorithm, \emph{SVMRank}, is limited.
Also, candidates of AutoScore and SVMRank are deemed more fluent than those of Top-1 and Random (2.33 vs.\ 2.29 and 2.26). In terms of argument quality, the results deviate from the automatic evaluation (Table~\ref{tab:exp1-auto}), showing marginally higher scores for SVMRank and Top-1.
Further analysis revealed that AutoScore and SVMRank agreed on the optimal candidate in 35\% of the cases, partially explaining their close scores. Although SVMRank achieved high scores across the three quality metrics, we note that the annotators preferred candidates scores generated by AutoScore, highlighting the importance of more diverse revision changes reflected by lower meaning preservation scores.

Overall, our findings suggest that using candidate selection approaches that incorporate quality assessments (i.e., {AutoScore} and {SVMRank}) leads to candidates of higher fluency and argument quality while preserving the meaning of the original claim. In addition to Figure~\ref{example_revisions}, examples of automatically-generated optimized claims can be found in the appendix.

\subsection{Performance with Context Integration}

Table \ref{tab:ablation_res} shows the semantic similarity of claims optimized by our approach and context information, depending on the context given. The results reveal slight improvements when conditioning the model on the previous claim (e.g., 60.3 vs.\ 59.4 BLEU). To check whether this led to improved claims, two authors of the paper compared 600 claims generated with and without the use of the previous claim in terms of (a)~which claim seems better overall and (b) which seems more grounded. We found that using the previous claim as context improved~quality in 12\% of the cases and lowered it in 1\% only, while leading to more grounded claims in 36\%.

\paragraph{Qualitative Analysis}

Our manual inspection of a claim sample revealed the following insights:

First, conditioning on context reduces the number of erroneous specifications, particularly for very short claims with up to 10 words. This seems intuitive, as such claims often convey little information about the topic of the debate, making inaccurate changes without additional context likely.

\begin{table}[t]
    \small
    \renewcommand{\arraystretch}{0.95}
    \centering
    \setlength{\tabcolsep}{4pt}
    \begin{tabular}{lrrrcrr}
        \toprule
         \textbf{Context}  &
        \textbf{BLEU} &
        \multicolumn{1}{l}{\bf Original} &
        \multicolumn{1}{l}{\bf Previous} &
        \multicolumn{1}{l}{\bf Topic} \\

        \midrule

       Claim only & 59.4           & 0.95 & 0.55    & 0.55    \\
       + Previous Claim   & \textbf{60.3}      & 0.95    & \textbf{0.57}&   \textbf{0.57}       \\
       + Debate Topic & 60.0                    & 0.95     & 0.55&  0.55   \\
        \midrule
        Human-Baseline & 100.0                    & 0.94          & 0.55  &  0.55 \\
        \bottomrule
    \end{tabular}
    \caption{BLEU and semantic similarity score with respect to the {\em original} claim, the debate's {\em previous} claim, and its {\em topic} of BART+AutoScore, depending on the context given for the 600 test samples.}
    \label{tab:ablation_res}
\end{table}

\begin{table*}[t]
    \small
    \renewcommand{\arraystretch}{0.6}
    \centering
    \setlength{\tabcolsep}{3pt}
    \begin{tabular}{llp{0.53\textwidth}rrr}
        \toprule
        \bf \# &
        \bf Optimization &
        \bf Description of the Type &
        \multicolumn{1}{l}{\bf Clarification} &
        \multicolumn{1}{l}{\bf Grammar} &
        \multicolumn{1}{l}{\bf Links} \\

        \midrule

        1 & Specification &
        Specifying or explaining a given fact or meaning (of the argument) by adding an example or discussion without adding new information. &
        58 &
        1 &
        -- \\[1.5em]
        % \addlinespace
        2 & Simplification &
        Removing information or simplifying the sentence structure, e.g., with the intent to reduce the complexity or breadth of the claim. &
        43 &
        -- &
        -- \\[1.5em]
        % \addlinespace

        3 & Reframing &
        Paraphrasing or rephrasing a claim, e.g., with the intent to specify or generalize the claim, or to add clarity. &
        29 &
        -- &
        -- \\[1.5em]
        % \addlinespace

       4 &  Elaboration &
        Extending the claim by more information or adding a fact with the intent to make the claim more self-contained, sound, or stronger. &
        23 &
        -- &
        -- \\[1.5em]
        % \addlinespace

        5 & Corroboration &
        Adding, editing, or removing evidence in the form of links that provide supporting information or external resources to the claim. &
        8 &
        -- &
        153 \\[1.5em]
        % \addlinespace

        6 & Neutralization &
        Rewriting a claim using a more encyclopedic or neutral tone, e.g., with the intent to remove bias or biased language. &
        7 &
        -- &
        -- \\[1.5em]
        % \addlinespace

        7 & Disambiguation &
        Reducing ambiguity, e.g., replacing pronouns by concepts mentioned before in the debate, or replacing acronyms with what they stand for. &
        7 &
        -- &
        1 \\[1.5em]
        % \addlinespace

        8 & Copy editing &
        Improving the  grammar, spelling, tone, or punctuation of a claim, without changing the main point or meaning. &
        41 &
        200 &
        52 \\

        \bottomrule
    \end{tabular}
    \caption{Descriptions of the eight claim optimization types identified in the 600 test pairs. The right columns show the count of claims per type for each of the three intention labels from \newcite{skitalinskaya2021}: {\em clarification}, typo/{\em grammar} correction, and correcting/adding {\em links}. Note, that a revision may be assigned to multiple categories.}
    \label{tab:edit-categories}
\end{table*}

Next, Kialo revisions often adhere to the following form: A claim introduces a statement and/or supporting facts, followed by a conclusion. This pattern was frequently mimicked by our approach. Yet, in some cases, it added a follow-up sentence repeating the original claim in different wording or generated conclusions containing fallacious or unsound phrases contradicting the original claim in others. Modeling context mitigated this issue.

Finally, we found that models conditioned on different contexts sometimes generated candidates optimized in different regards, whereas a truly optimal candidate would be a fusion of both suggestions.

    \section{Analysis}
\label{sec:analysis}

To explore the nature of claim optimization and the capabilities of our approach, this section reports on (a)~what types of optimizations exist, (b)~how well our approach can operationalize these, and (c)~how well it generalizes to non-argumentative domains.

\subsection{Taxonomy of Optimization Types}

To understand the relationship between optimizations found in the data and the underlying revision intentions, two authors of this paper inspected 600 claim revisions of the test set. Opposed to actions, intentions describe the goal of an edit (e.g., making a text easier to read) rather than referring to specific changes(e.g., paraphrasing or adding punctuation).\,
We build on ideas of \citet{yang_identifying_2017} who provide a taxonomy of revision intentions in Wikipedia texts. Claims usually do not come from encyclopedias, but from debate types or from monological arguments, as in essays \cite{persing:2015}. Therefore, we adapt the terminology of \citet{yang_identifying_2017} to gear it more towards argumentative texts.

As a result of a joint discussion of various sample pairs, we decided to distinguish eight optimization types, as presented in Table~\ref{tab:edit-categories}. Both authors then annotated all 600 test pairs for these types, which led to only 29 disagreement cases, meaning a high agreement of 0.89 in terms of Cohen's $\kappa$. These cases were resolved by both annotators together.%
\footnote{We acknowledge that there is potential bias inherent in self-annotation. However, we would like to point out that no knowledge about the test set was used to develop the approach presented in Section~\ref{sec:approach}.}

Table~\ref{tab:edit-categories} also shows cooccurrences of the types and intention labels. \emph{Typo/grammar correction} and \emph{correcting/adding links} align well with \emph{copy editing} and \emph{corroboration} respectively. In contrast, clarification is broken into more fine-grained types, where {\em specification} seems most common with 58 cases, followed by {\em simplification} and {\em reframing}. Examples of each type are found in the appendix.\,

We point out that the eight types are not exhaustive for all possible claim quality optimizations, but rather provide insights into the semantic and discourse-related phenomena observed in the data.
We see them as complementary to the argument quality taxonomy of \citet{wachsmuth-etal-2017-computat} as ways to improve the delivery-related quality dimensions: {\em clarity}, {\em appropriateness}, and {\em arrangement}.

\begin{table}[t]
\centering
\renewcommand{\arraystretch}{0.95}
\setlength{\tabcolsep}{3pt}
\small
\begin{tabular}{l@{\hspace*{-0.25cm}}rrrrr}
\toprule
\bf Type		& \bf Human	& \bf Approach	& \bf Better	& \bf Same & \bf Worse	\\
\midrule
Specification 		& 59			& 152		& 65\%		& 19\%		& 16\%	\\ 
Simplification 		& 43			& 18 			& 61\%		& 28\%		&11\%	\\ 
Reframing			& 29 			& 21			& 62\%		& 33\%		&5\%	\\ 
Elaboration 		& 23			& 55			& 62\%		& 18\%		&	20\%\\ 
Corroboration 		& 161		& 38			& 53\%		& 23\%		& 24\%	\\ 
Neutralization		& 7			& 0			& --\phantom{\%} & --\phantom{\%}		& --\phantom{\%}\\ 
Disambiguation 	& 8			& 8			& 63\%		& 25\%		& 12\%	\\ 
Copy editing		& 293		& 301		& 59\%		& 26\%		& 15\%	\\
\midrule
\textbf{Overall}		& 623		& 593 & 60\%		& 24\%		&16\%	\\ 
\bottomrule
\end{tabular}
\caption{Manual analysis: Comparison of the human-optimized claims of all 600 test cases (some have multiple) and of the claims optimized by BART+AutoScore (15 claims were unchanged). The three right columns show the ratio of optimized claims judged {\em better}, {\em same}, or {\em worse} than the original in terms of overall quality.}
\label{tab:category-success}
\end{table}

\subsection{Performance across Optimization Types}

To enable comparison between the human optimizations and automatically generated outputs, two authors of the paper labeled 600 optimized claims with the types defined in Table~\ref{tab:edit-categories}. Due to resource constrains only the best performing approach, BART+AutoScore, was considered. 
Overall, our approach generates better claims in 60\% of the cases, while 84\% remain at least of similar quality.

Most noteworthily, we observe that our approach performs optimizations of the type {\em specification} 2.5 times as often as humans, and more than double as many {\em elaboration} revisions (55 vs.\ 23). In contrast, it adds, edits, or removes evidence in the form of links ({\em corroboration}) four times less often than humans. The model also made fewer {\em simplifications} (18 vs.\ 43) and no {\em neutralization} edits, which may be due to data imbalance regarding such types.

In terms of average quality, {\em specification} (65\%) and {\em disambiguation} edits\,(63\%) most often lead~to improvements, but the eight types appear rather balanced in this regard. The Jaccard similarity~score between optimizations performed by humans and our approach is $0.37$, mostly agreeing on copy~edits\,(178 cases) and corroboration\,(22 cases). Gi\-ven such low overlap, future work should consider conditioning models to generate specific optimizations.

\subsection{Performance across Revision Domains}
\label{subsec:domains}

Lastly, we examine whether our approach, along with the chosen text quality metrics, applies to texts from other domains. We consider two datasets: \emph{WikiHow} \cite{anthonio-roth-2020-learn}, containing revisions of instructional texts, and \emph{IteraTeR} \cite{du-etal-2022-understanding-iterative}, containing revisions of various formal texts, such as encyclopedia entries, news, and scientific papers. For our experiments, we use the provided document-level splits, and sample 1000 revision pairs pseudo-randomly as a final test set.

Table \ref{tab:exp1-auto-ext} shows automatic evaluation results. In both cases, \emph{BART+Autoscore} leads to higher SARI scores (48.5 vs.\ 41.3 for WikiHow, 38.6 vs.\ 37.0 for IteraTeR), and notably reduces the number of cases where the models failed to revise the input (0.08 vs.\ 0.50 for WikiHow).\,The reported \emph{BART+Top1} model represents the approach~of \citet{du-etal-2022-understanding-iterative}, indicating that our approach and its text quality metrics achieve state-of-the-art performance with systematic improvements across domains, when generating optimized content. However, as different domains of text have different goals, different notions of quality, and, subsequently, different revision types performed, integrating domain-specific quality metrics may further improve performance. We leave this for future work.

\begin{table}[t]
    \renewcommand{\arraystretch}{0.95}
    \setlength{\tabcolsep}{2pt}
    \centering
    \small
    \begin{tabular}{l@{\hspace*{-0.5em}}rrrcr}
        \toprule
        \bf Approach &
        \multicolumn{1}{l}{\bf BLEU} &
        \multicolumn{1}{r}{\bf RouL} &
        \multicolumn{1}{l}{\bf SARI} &
        \multicolumn{1}{r}{\bf NoEd$\downarrow$} &
        \multicolumn{1}{c}{\bf ExM}  \\

        \midrule

\bf WikiHow Dataset \\
 \quad Unedited & 65.7 & 0.85 & 28.4 & 1.00&  0.00\% \\

 \quad BART + Top-1      	&  \bf 64.7  & \bf 0.83 & 41.3 & 0.50 & 13.0\% \\

  \quad	BART + AutoScore    		&   61.8 & 0.80	& \bf 48.5 & \bf 0.08 & \bf 16.0\%  \\[0.5ex]

\multicolumn{2} {l}{\bf IteraTeR Dataset} \\

\quad Unedited  & 74.0 & 0.86 & 28.6& 1.00& 0.00\%  \\

 \quad BART + Top-1          		&    \bf 68.9 & \bf 0.83 & 37.0 & 0.07 & 0.00\%  \\

  \quad	BART + AutoScore    		&   64.8 & 0.80 & \bf 38.6 & \bf 0.02 & 0.00\% \\

\bottomrule
\end{tabular}
\caption{Automatic evaluation: Performance of candidate selection strategies on data from other domains, in terms of BLEU, Rouge-L, SARI, ratio of unedited samples, and ratio of exact matches to target reference.}
\label{tab:exp1-auto-ext}
\end{table}

    \section{Conclusion}

With this paper, we work towards the next level of computational argument quality research, namely, to not only {\em assess} but also to {\em optimize} argumentative text.  Applications include suggesting improvements in writing support and automatic phrasing in debating systems. We presented an approach~that generates multiple candidate claim optimizations and then selects the best one using\,various quality metrics. In experiments, combining fine-tuned BART with such candidate selection improved 60\% of the claims from online debates, outperforming several baseline models and candidate selection strategies. We showcased generalization capabilities on two out-of-domain datasets, but we also found some
claim optimization types hard to automate.\,

In future work, we seek to examine whether recent large language models (e.g., Alpaca) and end-to-end models (where generation and candidate selection are learned jointly) can further optimize the quality of claims. As our approach so far relies on the availability of large claim revision corpora and language models, techniques for low-resource scenarios and languages should be explored to make claim optimization more widely applicable.

 \section*{Acknowledgments} This work was partially funded by the Deutsche Forschungsgemeinschaft (DFG, German Research Foundation) under project number 374666841, SFB 1342. 
    \section*{Ethical Considerations}

This work contributes to the task of argumentative text editing, namely we explore how to revise claims automatically in order to optimize their quality. While our work may also improve downstream task performance on other tasks, it is mainly intended to support humans in scenarios, such as the creation and moderation of content on online debate platforms as well as the improvement of arguments generated or retrieved by other systems. In particular, the presented approach is meant to help users by showing examples of how to further optimize their claims in relation to a certain debate topic, so they can deliver their messages effectively and hone their writing skills.

However, our generation approach still comes with limitations and may favor revision patterns over others in unpredictable ways, both of which might raise ethical concerns. For example, it may occasionally produce false claims based on untrue or non-existent facts. We think, humans should be able to identify such cases in light of the available context though, as long as the improvements remain suggestions and do not happen fully automatically, as intended.

The presented technology might further be subject to intentional misuse. A word processing software, for example, could be conditioned to automatically detect and adapt claims made by the user in subtle ways that favors political or social views of the software provider. Such misuse might then not only change the intended message of the text, but also influence or even change the views of the user \cite{jakesch-2023}.

In a different scenario, online services, such as social media platforms or review portals, might change posted claims (e.g. social media posts, online reviews) to personalize them and increase user engagement or revenue. These changes might not only negatively affect the posting, but also the visiting user.

While it is hard to prevent such misuse, we think that the described scenarios are fairly unlikely, as such changes tend to be noticed by the online community quickly. Furthermore, the presented architecture and training procedure would require notable adaptations to produce such high-quality revisions.

An aspect that remains unexplored in this work is the ability of the presented approaches to work with variations of the English language, such as African-American English, mainly due to the lack of available data. In this regard, the approach might unfairly disadvantage or favor particular language varieties and dialects, potentially inducing social bias and harm if applied in public scenarios. We encourage researchers and practitioners to stay alert for such cases and to choose training data with care for various social groups.

 Finally, our work included the labeling of generated candidate claims on a crowdsourcing platform. As detailed in Section \ref{sec:experiments}, we compensated MTurk workers \$13 per hour, complying with minimum wage standards in most countries at the time of conducting the experiment.

    \bibliography{inlg23-claim-revision-generation-lit}
    \bibliographystyle{acl_natbib}

    \appendix
    \clearpage
\appendix
\section{Implementation and Training Details}
\label{app:implementation}

\subsection{Candidate Generation Models}

For generation, we use the pre-trained BART model implemented in the fairseq library. The library and pre-trained models are BSD-licensed. We use the BART-large checkpoint (400M parameters) and further finetune the model for 10 epochs on 2 RTX 2080Ti GPUs. We use the same parameters as suggested in the fine-tuning of BART for the CNN-DM summarization task by fairseq
and set MAX-TOKENS to 1024. The training time is 100-140 minutes, depending on the chosen setup (with or without context information).

During inference, we generate candidates using a top-k random sampling scheme \cite{fan-etal-2018-hierarchical}  with the following parameters: length penalty is set to 1.0, n-grams of size 3 can only be repeated once, temperature is set to 0.7, while the minimum and maximum length of the sequence to be generated are 7 and 256 accordingly.

\subsection{Quality Assessment Models}

For the automatic assessment of fluency and argument quality, we use the bert-base-cased pre-trained BERT version, as implemented in the huggingface library. The library and pre-trained models have the Apache License 2.0. We finetune the model for two epochs and use the parameters suggested in \citet{skitalinskaya2021}. The accuracy of the trained model for fluency obtained on the train/dev/test split suggested by the authors \cite{toutanova-etal-2016-dataset} is 77.4  and 75.5 for argument quality. 

For labeling the missing or unassigned revision types, we use the same bert-base-cased pre-trained BERT model, but in a multi-label setup, where we consider the following 6 classes: claim clarification, typo or grammar correction, correcting or adding links, changing the meaning of the claim, splitting the claim, and merging claims. We fine-tune the model for two epochs using the Adam optimizer with a learning rate of 1e-5 and achieve a weighted F1-score of 0.81.

\section{Alternative Generation Models}
\label{sec:alternatives}

For comparison,\,we provide two additional baseline Seq2Seq model architectures,\,which help identify the complexity of the model needed for the task:

\textbf{LSTM.} Our first baseline is a popular LSTM variant introduced by \citet{wiseman2016}.  We use the \textit{lstm\_wiseman\_iwslt\_de\_e} architecture, which is a two-layer encoder and decoder LSTM, each with 256 hidden units, and dropout with a rate of 0.1 between LSTM layers.

\textbf{Transformer.} The second model is based on the work of \citet{vaswani2017}.  We use the \textit{transformer\_iwslt\_de\_en} architecture, a 6-layer encoder and decoder with 512-dimensional embeddings, 1024 for inner-layers, and four self-attention heads.

Tables \ref{tab:exp1-appendix} and \ref{tab:exp1-details-appendix} compare the automatic evaluation scores of all generation-content selection combinations. 

\begin{table}[t]
    \renewcommand{\arraystretch}{0.95}
    \setlength{\tabcolsep}{1.7pt}
    \centering
    \small
\begin{tabular}{llrrrcr}
        \toprule
        \bf Model & \bf Strategy& 
        \multicolumn{1}{l}{\bf BLEU} &
        \multicolumn{1}{r}{\bf RouL} &
        \multicolumn{1}{l}{\bf SARI} &
        \multicolumn{1}{r}{\bf NoEd$\downarrow$} &
        \multicolumn{1}{c}{\bf ExM}  \\

        \midrule
BART & Top-1          		& 64.0	    &  0.83 		& 39.7	& 0.31    & 7.8\%     \\

 & Random     		& 62.6      & 0.83 		& 38.7 	& 0.28    & 6.8\%      	\\
  
 & SVMRank    		& 55.7      & 0.76 		& 38.8  & 0.03  & 4.5\%        	\\
& AutoScore		& 59.4      & 0.80 	    &   43.7 	& 0.02 & 8.3\%	\\
\addlinespace
Trans- 		& Top-1 			& 43.6		& 0.64  		& 0.30          	& 0.12 & 0.8\%         \\
former         	& Random 		& 42.4         	& 0.63  		& 0.30          	& 0.13 & 1.0\%         \\
         		& SVMRank		& 41.8         	& 0.63  		& 0.31          	& 0.10 & 1.2\%       \\
         		& AutoScore		& 40.5          	& 0.62	 	& 0.30          	& 0.10 & 1.3\% \\
\addlinespace
LSTM  		&  Top-1   			& 36.2		& 0.56 		& 0.28          	& 0.10 & 0.3\%       \\
         		&  Random   		& 36.0          	& 0.56 		& 0.28          	& 0.10 & 0.3\%          \\
         		&  SVMRank   		& 36.2		   & 0.56 		& 0.29		& 0.10 & 1.0\%      \\
         		&  AutoScore	& 34.1          	& 0.52 		& 0.28          	& 0.10 & 1.0\%\\
\bottomrule
\end{tabular}
\caption{Automatic evaluation: Results for each combination of generation model and candidate selection strategy on the 600 test samples,  in comparison to the human revisions: BLEU (0-100), ROUGE-L (RouL), SARI, ratio of unedited samples (NoEd), \% of exact matches to target reference (ExM).}
\label{tab:exp1-appendix}
\end{table}
\begin{table}[t]
    \renewcommand{\arraystretch}{0.95}
    \setlength{\tabcolsep}{1.7pt}
    \centering
    \small
    \begin{tabular}{llrrcc}
        \toprule
        \bf Model &
        \bf Strategy  & 
        \multicolumn{1}{r}{\bf Fluency} &
        \multicolumn{1}{l}{\bf Meaning} &
        \multicolumn{1}{r}{\bf Argument} & \multicolumn{1}{r}{\bf Average}  \\

        \midrule

BART		& Top-1         & 0.73  		& 0.97	& 0.65     & 0.78     \\
        			& Random     	            	& 0.72 		& 0.97 	& 0.68  & 0.79        	\\
        			& SVMRank    	      	& 0.72 		& 0.94          	& 0.76   & 0.81       	\\
        			& AutoScore	       	& 0.83 	& 0.95 		& 0.86 & \bf 0.88	\\
\addlinespace        			
Trans-		& Top-1         & 0.44  		& 0.76	& 0.40     & 0.53     \\
former & Random     	            	& 0.41 		& 0.76 	& 0.38  & 0.52        	\\
& SVMRank    	      	& 0.50 		& 0.76          	& 0.45   & 0.57       	\\
& AutoScore	       	& 0.68 	& 0.75 		& 0.61 & \bf 0.68	\\
\addlinespace
LSTM		& Top-1         & 0.27  		& 0.68	& 0.31     & 0.42     \\
& Random     	            	& 0.27 		& 0.68 	& 0.31  & 0.42        	\\
& SVMRank    	      	& 0.29 		& 0.69          	& 0.31   & 0.43       	\\
& AutoScore	       	& 0.52 	& 0.65 		& 0.53 & \bf 0.57	\\
\midrule
Human   		&  	       	& 0.72 		& 0.94          	& 0.74  & 0.80    \\
\bottomrule
\end{tabular}
\caption{Results for each combination of generation model and candidate selection strategy on the 600 test samples, in comparison to the human revisions based on three quality metrics: fluency, meaning preservation and argument quality.}
\label{tab:exp1-details-appendix}
\end{table}

\subsection{Automatic Evaluation}

We use the following python packages and scripts to perform automatic evaluations: nltk (BLEU \cite{papineni-etal-2002-bleu}), rouge-score (ROUGE \cite{lin-2004-rouge}),  \href{https://github.com/cocoxu/simplification/SARI.py}{https://github.com/cocoxu/simplification/ SARI.py} (SARI \cite{xu-etal-2016-optimizing})

\section{Claim Optimization Examples}

For all eight optimization categories, we provide one or more examples illustrating each action in Table \ref{tab:category-examples}.

\begin{table*}[]
\small
\setstcolor{red}
\begin{tabular}{p{0.12\linewidth}p{0.80\linewidth}}
\toprule
  \textbf{Type} &
  \textbf{Examples} \\ \midrule
  
  Specification &   Nipples are the openings of female-only {\leavevmode\color{forestgreen}exocrene} glands that can have abnormal [secretions] \textless{}LINK\textgreater\ during any time of life, get erected by cold stimulation or sexual excitement (much more visibly than in men), get lumps or bumps and change color and size of areola during the menstrual cycle or pregnancy, so their display can break [personal space] \textless{}LINK\textgreater\ and privacy (which is stressful), affect public sensibilities and also be a [window] \textless{}LINK\textgreater\ for infections, allergies, and irritation. \\
  \\
  & The idea behind laws{\leavevmode\color{forestgreen}, such as limiting the amount of guns,} is to reduce the need to defend yourself from a gun or rapist. \\
  \\
  & It is very common for governments to actively make certain forms of healthcare [harder for minority groups to access] \textless{}LINK\textgreater . {\leavevmode\color{forestgreen} They could also, therefore, make cloning technology hard to access.} \\ \midrule
  
  Simplification & Very complex{\leavevmode\color{red}\st{, cognitively meaningful behavior such as}} {\leavevmode\color{forestgreen} behaviours like creating} art are evidence of free will{\leavevmode\color{red}\st{, because they exhibit the same lack of predictability as stochastic systems, but are intelligible and articulate clearly via recognizable vehicles}}. \\ \midrule
  
  Reframing & It reduces the oversight of the BaFin and thus increases {\leavevmode\color{red}\st{the risk of financial crisis}} {\leavevmode\color{forestgreen} market failures}. \\ \midrule
  
  Elaboration & 
  It takes 2-4 weeks for HIV {\leavevmode\color{forestgreen} to }present any symptom. The incubation period risk {\leavevmode\color{red}\st{can't be ruled out for}}  {\leavevmode\color{forestgreen}is higher for a} member of high risk group, {\leavevmode\color{red}\st{effectively and timely}} {\leavevmode\color{forestgreen}even though member of a low risk group is not completely safe. The decision is based on the overall risk, not on individual level.} \\ \midrule
  Corroboration & [Person-based predictive policing technologies] {\leavevmode\color{forestgreen}\textless{}LINK\textgreater} - that focus on predicting who is likely to commit crime rather than where is it likely to occur - violate the [presumption of innocence.] {\leavevmode\color{forestgreen}\textless{}LINK\textgreater}. \\ \midrule
  
  Neutralization & Biden {\leavevmode\color{red}\st{does not}} {\leavevmode\color{forestgreen} lacks the} support {\leavevmode\color{red}\st{ or agree with several key issues that are important to liberal voters.}} {\leavevmode\color{forestgreen} of many liberal voting groups due to his stance on key issues concerning them.}
  \\ \midrule
  
  Disambiguation & The USSR had [passed legislation] \textless{}LINK\textgreater\ to gradually eliminate religious belief within its borders. However the death penalty was more used in USSR than in Russia. {\leavevmode\color{red}\st{It}} {\leavevmode\color{forestgreen}USSR} had 2000 [death penalties] \textless{}LINK\textgreater\ per year {\leavevmode\color{forestgreen}in the 1980s} whereas pre USSR Russia had [banned the death penalty] \textless{}LINK\textgreater\ in 1917 and almost never carried it out in the decades before that. \\ 
  \\
  & {\leavevmode\color{red}\st{SRM}} {\leavevmode\color{forestgreen}Solar geoengineering} merely serves as a "technological fix" (Weinberg).[harvard.edu] \textless{}LINK\textgreater \\ \midrule
  
  Copy Editing & Women are experiencing record {\leavevmode\color{red}\st{level}} {\leavevmode\color{forestgreen}levels} of success in primaries.\\
 \bottomrule
\end{tabular}
\caption{Illustrative examples of optimization types identified in the paper. The green font denotes additions and the striked out red font denotes the removal of text snippets.}
\label{tab:category-examples}
\end{table*}

\section{Manual Quality Assessment Guidelines}
Figure \ref{fig:annotation-guidelines} shows the annotation guidelines for the Amazon Mechanical Turk study.

\begin{figure*}
    \centering
    \begin{mdframed}
        {\Large\bf Instructions}
        \vspace{2mm}

        {\small
            In this task, your goal is to identify whether a claim has been successfully improved, without changing the overall meaning of the text.
    
            Each task contains a set of pairs, where one claim is the "original claim," and the other an optimized candidate. Each of these pairs have the same original text, but different candidate optimizations.
            
            Please rate each candidate along the following three perspectives: argument quality, fluency and semantic similarity. And, finally, please, rank all candidates relative to each other in terms of overall quality.
        }
        \vspace{3mm}

        {\bf Argument Quality}
        \vspace{1mm}
        
        {\small
            \textbf{Scale (1-5)}: 1 (notably worse than original), 2 (slightly worse), 3 (same as original), 4 (slightly improved), 5 (notably improved)
            \vspace{1mm}
    
            Does the optimized claim improve the argument quality compared to the original claim? Relevant changes include, but are not limited to:
            \begin{itemize}
            \setlength\itemsep{-0.35em}
                \item further specifying or explaining an existing fact or meaning
                \item removing information or simplifying the sentence structure with the intent to reduce the complexity or breadth of the claim
                \item rephrasing a claim with the intent to specify or generalize the claim, or to add clarity
                \item adding  (substantive)  new  content  or  information  to  the  claim  or inserting an additional fact with the intent of making it more self-contained, more sound or stronger
                \item adding, editing or removing evidence in the form of links that provide supporting information or external resources to the claim
                \item removal of bias or biased language
                \item removal uncertainty.  e.g.  by replacing pronouns referring to concepts that have been mentioned in other claims of the debate, or by replacing acronyms with what they stand for
                \item improving the grammar, spelling, tone, or punctuation of a claim
            \end{itemize}
        }
        \vspace{1mm}

        {\bf Meaning}
        \vspace{1mm}
        
        {\small
            \textbf{Scale (1-5)}: 1 (entirely different), 2 (substantial differences), 3 (moderate differences), 4 (minor differences), 5 (identical)
            \vspace{1mm}
    
            Does the transformed claim still have the same overall meaning as the original? It is OK if extra information is added, as long as it doesn't change the underlying people, events, and objects described in the sentence. You should also not strongly penalize for meaning transformations which aim to generalize or specify some aspects of the claim.
        }
        \vspace{3mm}

        {\bf Fluency}
        \vspace{1mm}
        
        {\small
            \textbf{Scale (1-3)}: 1 (major errors, disfluent), 2 (minor errors), 3 (fluent)
            \vspace{1mm}
    
            Is this sentence fluent English and does it make sense? 
            
            \vspace{3mm}
            Examples with explanations can be found here.
        }
    \end{mdframed}
    \caption{Guidelines provided to the crowdworkers to annotate automatically generated claim optimizations.}
    \label{fig:annotation-guidelines}
\end{figure*}

\section{System Outputs}

Table \ref{tab:ranking-examples} provides examples of candidates selected by different content selection strategies along with human references illustrating common patterns found in the results. Table \ref{tab:context-examples} provides examples of candidates generated with and without utilizing context knowledge with insertions and deletions being highlighted in green and red fonts accordingly.

\begin{table*}[]
\small
\setstcolor{red}
\begin{tabular}{p{0.07\linewidth}p{0.87\linewidth}}
\toprule
  \textbf{Strategy} &
  \textbf{Candidate} \\ \midrule
  
  Original &   Easy to cheat, and procastinate \\
  Top-1 & {\leavevmode\color{forestgreen} It is} easy to cheat and procastinate {\leavevmode\color{forestgreen}.}\\
  SVMRank & \leavevmode\color{forestgreen}{The} \color{black}{easy to cheat, and procastinate} \\
  AutoScore & \leavevmode\color{forestgreen}{The UBI is} \color{black}{easy to cheat and} \color{forestgreen}{easy to manipulate.}\\
  Human & {\leavevmode\color{forestgreen} Flexible schedules can easily increase procrastination}
  \\ \midrule

Original & Many games often attempt to depict the norms and attitudes of the real world. \\
  Top-1 &
  Many games often attempt to depict the norms and attitudes of the real world. 
  \\

  SVMRank &
  Many games often attempt to depict the norms and attitudes of the real world, \color{forestgreen}{and thus are likely to be accurate and relevant to what people are thinking and feeling in the real-world. }
  \\

  AutoScore &
  Many games often attempt to depict the norms and attitudes of the real world, \color{forestgreen}{in order to give players a greater understanding of the world around them.}  \\
  Human & Many games often attempt to depict the norms and attitudes of the real world\color{forestgreen}{; gamers are likely to see social attitudes espoused in these games on issues of gender equality as indicative of progress in the real world}.
  \\ \midrule
 Original & 
 Owners of PWA take home {[}more profit{]} \textless{}LINK\textgreater\ from their sales than mobile apps.  \\

  Top-1 &
  Owners of PWA take home {[}more profit{]} \textless{}LINK\textgreater\ from their sales than mobile apps. 
   \\

  SVMRank &
  Owners of PWA take home {[}more profit{]} \textless{}LINK\textgreater\ from their sales than mobile apps. \color{forestgreen}{This allows them to reinvest in their business and increase their profits.} 
   \\ 

  AutoScore &
  Owners of PWA take home {[}more profit{]} \textless{}LINK\textgreater\ from their sales than \color{forestgreen}{those of} mobile apps. 
   \\
   Human & Owners of \color{forestgreen}{PWAs} \color{black}{ take home [more profit] \textless{}LINK\textgreater\ from their sales than mobile apps.} \\
   
   \midrule
Original & Scientists found out that Saturn's moon {[}Enceladus{]} \textless{}LINK\textgreater\ houses a giant, global ocean beneath its icy outer shell.  \\
  Top-1  &
  Scientists found out that Saturn's moon {[}Enceladus{]} \textless{}LINK\textgreater\ houses a giant, global ocean beneath its icy outer shell. 
  \\

  SVMRank &
  Scientists found out that Saturn's moon {[}Enceladus{]} \textless{}LINK\textgreater\ houses a giant, global ocean beneath its icy outer shell. \color{forestgreen}{This suggests that this icy, water-filled place also houses other life forms, such as asteroids and comets.} 
  \\
  AutoScore &
  Scientists found out that Saturn's moon {[}Enceladus{]} \textless{}LINK\textgreater\ houses a giant, global ocean beneath its icy outer shell, \color{forestgreen}{suggesting that the moon may hold life beyond the surface.} \\
  Human & Scientists found \color{red}{\st{out}} \color{black}{that Saturn's moon {[}Enceladus{]} \textless{}LINK\textgreater\ houses a giant, global ocean beneath its icy outer shell.}
  \\ \midrule
  
  Original & Responsibility requires conscious decisions and [free will] \textless{}LINK\textgreater. Acting according to governmental orders is far from that. \\
  Top-1 & Responsibility requires conscious decisions and [free will] \textless{}LINK\textgreater. Acting according to governmental orders is far from that. \\
  SVMRank & Responsibility requires conscious decisions and [free will] \textless{}LINK\textgreater. Acting according to governmental orders is far from that. \color{forestgreen}{Therefore, people are not responsible for their actions, and are therefore not morally responsible.} \\
  AutoScore & Responsibility requires conscious decisions and [free will] \textless{}LINK\textgreater. \color{forestgreen}{Acting according to governmental orders is far from that, as it is not a conscious decision but rather a result of an involuntary reaction.} \\
  Human & {\leavevmode\color{forestgreen}The coercion to vote eliminates the idea of responsibility, which} requires conscious decisions and [free will] \textless{}LINK\textgreater. Acting according to governmental orders is far from that.
  \\ \midrule
  
  Original & There was much progress, advancement and forward looking thinking in the 20th century in Europe across all disciplines. \\
  Top-1 & There was much progress, advancement and forward looking thinking in the 20th century in Europe across all disciplines. \color{forestgreen}{Therefore, it is logical to conclude that the world has changed in the last 150 years.} \\
  SVMRank & There was much progress, advancement and forward looking thinking in the 20th century in Europe across all disciplines\color{forestgreen}{, which shows the progress that was made during the century}. \\
  AutoScore & There was much progress, advancement\color{forestgreen}{,} \color{black}{and forward looking thinking in the 20th century in Europe across all disciplines.} \\ 
  Human & There was much progress, advancement and forward \color{red}{\st{looking}} \color{black}{ thinking in the 20th century in Europe across all disciplines.}\\ \midrule
  Original & A [catch-22] \textless{}LINK\textgreater\ situation currently exists in regards to researching the medicinal applications of some illegal drugs, due to the laws surrounding how they are categorised ([p. 12] \textless{}LINK\textgreater). \\
  SVMRank (Top-1) & A [catch-22] \textless{}LINK\textgreater\ situation currently exists in regards to researching the medicinal applications of some illegal drugs, due to the laws surrounding how they are categorised ([p. 12] \textless{}LINK\textgreater). {\color{forestgreen}This prevents researchers from knowing how to fully assess the medicinal potential of drugs in a manner that is best suited for the patients who need it.} \\
  AutoScore & A [catch-22] \textless{}LINK\textgreater\ situation currently exists in regards to researching the medicinal applications of some illegal drugs, due to the laws surrounding how they are categorised ([p. 12] \textless{}LINK\textgreater).  {\color{forestgreen}This prevents researchers from researching possible medicinal uses for illegal drugs.}\\
  Human & A [catch-22] \textless{}LINK\textgreater\ situation currently exists in  {\color{forestgreen}regard} to researching the medicinal applications of some illegal drugs, due to the laws surrounding how they are categorised ([p. 12] \textless{}LINK\textgreater).
  \\ \bottomrule
\end{tabular}
\caption{Examples of optimizations selected by different candidate selection strategies along with human references. The green font denotes additions and the striked out red font denotes the removal of text snippets.}
\label{tab:ranking-examples}
\end{table*}

\begin{table*}[t]
\small
\setstcolor{red}
\begin{tabular}{lp{0.85\linewidth}}
\toprule 
  \textbf{Control} &
  \textbf{Candidate} 
  \\
  \midrule

\rowcolor[gray]{0.95} Topic &  Should Men Be Involved And Trusted In Childcare? \\
\rowcolor[gray]{0.95} Previous & Perpetrators of child abuse are {[}way more likely{]} \textless{}LINK\textgreater\ to be the child's parents.\\ 
\rowcolor{orange} Original & Mostly the mothers, too.\\ 
\rowcolor{lightergreen}  Claim only &
 {\leavevmode\color{forestgreen} The} mothers {\leavevmode\color{forestgreen}are the most vulnerable. }\\
\rowcolor{lightergreen}+ Previous &
  {\leavevmode\color{forestgreen}The majority of perpetrators are }the mothers, too. \\
\rowcolor{lightergreen}+ Topic &
 {\leavevmode\color{forestgreen} It is mostly} mothers {\leavevmode\color{forestgreen}who are responsible for childcare. }
   \\ \midrule  
   
   \rowcolor[gray]{0.95} Topic &  Would Trump or Biden be a better President? \\
\rowcolor[gray]{0.95} Previous & The {[}Healthy Indiana Plan 2.0{]} \textless{}LINK\textgreater{}, initiated by Mike Pence, has provided access to healthcare to many people who could not previously afford it. \\ 
\rowcolor{orange} Original & The Healthy Indiana Plan has high penalties for non-payment of {[}premiums{]} \textless{}LINK\textgreater{}.\\
\rowcolor{lightergreen}Claim only & 
  The Healthy Indiana Plan has high penalties for non-payment of {[}premiums{]} \textless{}LINK\textgreater{}. {\leavevmode\color{forestgreen}This means that people with low income could be penalized for failure to pay.}
\\

\rowcolor{lightergreen}+ Previous &
  The Healthy Indiana Plan has high penalties for non-payment of {[}premiums{]} \textless{}LINK\textgreater\ {\leavevmode\color{forestgreen}in exchange for providing healthcare. }
   \\
\rowcolor{lightergreen} + Topic &
  The Healthy Indiana Plan has high penalties for non-payment of {[}premiums{]} \textless{}LINK\textgreater{},{\leavevmode\color{forestgreen} which were imposed in 2015. }
   \\ \midrule

\rowcolor[gray]{0.95} Topic &  Artificial Intelligence (AI): Should an artificial general intelligence be created? \\
\rowcolor[gray]{0.95} Previous & There is the chance an AGI might be able to think for itself and turn on its creators.\\ 
\rowcolor{orange}Original & AGI are susceptable.\\
\rowcolor{lightergreen}  Claim only &
  {\leavevmode\color{forestgreen}Many} AGI are susceptable {\leavevmode\color{forestgreen}to the danger of death}.  \\
\rowcolor{lightergreen}+ Previos &
  AGI are susceptible. {\leavevmode\color{forestgreen}They may be influenced by humans and by the biases of their creators.} \\
\rowcolor{lightergreen}+ Topic &
  AGI are susceptible{\leavevmode\color{forestgreen} to being hacked. } \\ \midrule

\rowcolor[gray]{0.95} Topic &  Who should the Conservative Party choose to be the next UK Prime Minister? \\
\rowcolor[gray]{0.95}  Previous &  {[}Jeremy Hunt{]} \textless{}LINK\textgreater\\ 
\rowcolor{orange} Original & Hunt is the best placed candidate to unify Parliament in order to get a Brexit deal passed.\\ 
\rowcolor{lightergreen} Claim only &  Hunt is the best placed candidate to unify Parliament {\leavevmode\color{forestgreen}and the public} in order to get a Brexit deal passed. \\
\rowcolor{lightergreen} + Previous &
  Hunt {\leavevmode\color{forestgreen}could use his position as the Prime Minister} to unify Parliament in order to get a Brexit deal passed. \\
\rowcolor{lightergreen} + Topic &
  Hunt is the best placed candidate to {\leavevmode\color{forestgreen}{[}unify Parliament{]} \textless{}LINK\textgreater} in order to get a Brexit deal passed \\ \midrule
  
  \rowcolor[gray]{0.95} Topic &  Should high-income countries take in refugees? \\
\rowcolor[gray]{0.95}  Previous & Taking in refugees will increase criminality within host countries.\\ 
\rowcolor{orange} Original & Differences in criminal activity are a product, in part, of childhood [socioeconomic] \textless{}LINK\textgreater\ conditions. This is exacerbated by the longer [path] \textless{}LINK\textgreater\ to employment faced by refugees compared to other communities. \\ 
\rowcolor{lightergreen} Claim only &  Differences in criminal activity are a product, in part, of childhood [socioeconomic] \textless{}LINK\textgreater\ conditions. This is exacerbated by the longer [path] \textless{}LINK\textgreater\ to employment faced by refugees compared to other communities{\leavevmode\color{forestgreen}, making them more likely to get involved in crime}. \\
\rowcolor{lightergreen} + Previous &
  Differences in criminal activity are a product, in part, of childhood [socioeconomic] \textless{}LINK\textgreater\ conditions. This is exacerbated by the longer [path] \textless{}LINK\textgreater\ to employment faced by refugees compared to other communities. {\leavevmode\color{forestgreen}This will not increase criminality.} \\
\rowcolor{lightergreen} + Topic &
  Differences in criminal activity are a product, in part, of childhood [socioeconomic] \textless{}LINK\textgreater\ conditions. This is exacerbated by the longer [path] \textless{}LINK\textgreater\ to employment faced by refugees compared to other communities{\leavevmode\color{forestgreen}, which make it harder to find a job}. \\ \midrule
  
    \rowcolor[gray]{0.95} Topic &  Mark Twain used the N-word in The Adventures of Huckleberry Finn. Should it be censored? \\
\rowcolor[gray]{0.95}  Previous & Changing the N-word would skip a piece of the \(linguistic\) past and thus everyday life. As a result, people could start to forget this part of history.\\ 
\rowcolor{orange} Original & In Huckleberry Finn, Twain captured the essence of "[everyday midwest American English] \textless{}LINK\textgreater". \\ 
\rowcolor{lightergreen} Claim only &  In Huckleberry Finn, Twain captured the essence of "[everyday midwest American English] \textless{}LINK\textgreater".{\leavevmode\color{forestgreen} This is a common trait of the American English language.} \\
\rowcolor{lightergreen} + Previous &
  In Huckleberry Finn, Twain captured the essence of "[everyday midwest American English] \textless{}LINK\textgreater"{\leavevmode\color{forestgreen} by using the N-word in everyday conversation}. \\
\rowcolor{lightergreen} + Topic &
  In Huckleberry Finn, Twain captured the essence of "[everyday midwest American English] \textless{}LINK\textgreater"{\leavevmode\color{forestgreen}, which is a language that is often used by people who do not share his values}. \\\bottomrule
\end{tabular}
\caption{Examples of different candidates generated by BART + AutoScore with and without context information. The green font denotes additions of text snippets.}
\label{tab:context-examples}
\end{table*}
\end{document}